\documentclass[10pt,journal,compsoc,x11names]{IEEEtran}
\usepackage{amsmath,amsfonts}
\usepackage{algorithmic}
\usepackage{algorithm}
\usepackage{array}
\usepackage[caption=false,font=normalsize,labelfont=sf,textfont=sf]{subfig}
\usepackage{textcomp}
\usepackage{stfloats}
\usepackage{url}
\usepackage{verbatim}
\usepackage{graphicx}
\usepackage{cite}
\usepackage{multirow}
\usepackage{hyperref}
\hypersetup{
    colorlinks=true,      
    linkcolor=blue,       
    citecolor=red,        
    filecolor=magenta,    
    urlcolor=blue        
}

\usepackage{forest}
\usetikzlibrary{shadows}
\usepackage{amsmath}

\usepackage{amssymb}
\usepackage{cleveref}

\definecolor{deepred}{rgb}{0.698,0.133,0.133}
\definecolor{blue}{rgb}{0,0,1}
\usepackage{tikz}
\usepackage{amssymb}

\definecolor{lightcoral}{rgb}{0.94, 0.5, 0.5}
\definecolor{lightgreen}{rgb}{0.56, 0.93, 0.56}
\definecolor{harvestgold}{rgb}{0.85, 0.57, 0.0}
\definecolor{brightlavender}{rgb}{0.75, 0.58, 0.89}
\definecolor{capri}{rgb}{0.0, 0.75, 1.0}
\definecolor{carminepink}{rgb}{0.92, 0.3, 0.26}
\definecolor{celadon}{rgb}{0.67, 0.88, 0.69}
\definecolor{darkpastelgreen}{rgb}{0.01, 0.75, 0.24}

\definecolor{deepred}{rgb}{0.698,0.133,0.133}
\definecolor{blue}{rgb}{0,0,1}
\definecolor{lightergray}{RGB}{230,230,230}
\definecolor{DarkRed}{RGB}{130,25,0}
\definecolor{PurpleRed}{RGB}{204,0,102}
\definecolor{DarkGreen}{RGB}{30,130,30}
\definecolor{DarkBlue}{RGB}{0,0,250}
\definecolor{DarkYellow}{RGB}{255,128,0}
\definecolor{light-gray}{gray}{0.95}
\definecolor{lightgreen}{RGB}{231,255,219}
\definecolor{lightred}{RGB}{252,231,234}
\definecolor{lightyellow}{RGB}{250,253,191}
\definecolor{lightpurple}{RGB}{229,204,255}
\definecolor{lightblue}{RGB}{229,246,254}
\definecolor{value-modification}{RGB}{250, 217, 86}
\definecolor{digit-expansion}{RGB}{216, 194, 104}
\definecolor{integer-decimal-fraction}{RGB}{240, 133, 51}
\definecolor{semantic-paraphrasing}{RGB}{85, 157, 63}
\definecolor{complexity-increasing}{RGB}{58, 120, 175}
\definecolor{question-transformation}{RGB}{174, 205, 225}
\definecolor{interference-injection}{RGB}{255,204,229}
\definecolor{remove-constrain}{RGB}{204,204,255}
\definecolor{myGreen}{RGB}{127,210,85}
\definecolor{myOrange}{RGB}{242,154,66}
\definecolor{myYellow}{RGB}{247,223,65}
\definecolor{myRed}{RGB}{232,80,43}
\definecolor{myViolet}{RGB}{162,57,102}
\definecolor{myBlue}{HTML}{4686f3}
\definecolor{myYellowv2}{HTML}{E6C802}
\definecolor{myOrangev2}{HTML}{ED8E55}
\definecolor{MyGreenv2}{HTML}{009B55}
\definecolor{MyRedv2}{HTML}{c22f2f}

\definecolor{lightcoral}{rgb}{0.94, 0.5, 0.5}
\definecolor{lightgreen}{rgb}{0.56, 0.93, 0.56}
\definecolor{harvestgold}{rgb}{0.85, 0.57, 0.0}
\definecolor{brightlavender}{rgb}{0.75, 0.58, 0.89}
\definecolor{capri}{rgb}{0.0, 0.75, 1.0}
\definecolor{carminepink}{rgb}{0.92, 0.3, 0.26}
\definecolor{celadon}{rgb}{0.67, 0.88, 0.69}
\definecolor{darkpastelgreen}{rgb}{0.01, 0.75, 0.24}

\definecolor{deepred}{rgb}{0.698,0.133,0.133}
\definecolor{blue}{rgb}{0,0,1}

\crefformat{section}{\S#2#1#3} 
\crefformat{subsection}{\S#2#1#3}
\crefformat{subsubsection}{\S#2#1#3}

\usepackage{amsthm}
\usepackage{amsmath}%
\usepackage{MnSymbol}%

\makeatletter
\newtheorem*{rep@theorem}{\rep@title}
\newcommand{\newreptheorem}[2]{%
\newenvironment{rep#1}[1]{%
 \def\rep@title{#2 \ref{##1}}%
 \begin{rep@theorem}}%
 {\end{rep@theorem}}}
\makeatother

\newreptheorem{theorem}{Theorem}

\newreptheorem{lemma}{Lemma}

\newreptheorem{proposition}{proposition}

\usepackage{bm}

\usepackage{booktabs}
\usepackage{xcolor,pifont}
\usepackage{bbding}
\usepackage{colortbl}
\usepackage{bm}

\definecolor{LightBlue}{cmyk}{0.09, 0.03, 0.01, 0.0}

\usepackage[inline,shortlabels]{enumitem}

\usepackage{tikz,xcolor,hyperref}
\definecolor{lime}{HTML}{A6CE39}
\DeclareRobustCommand{\orcidicon}{%
    \begin{tikzpicture}
    \draw[lime, fill=lime] (0,0) 
    circle [radius=0.16] 
    node[white] {{\fontfamily{qag}\selectfont \tiny ID}};    \draw[white, fill=white] (-0.0625,0.095) 
    circle [radius=0.007];    \end{tikzpicture}
    \hspace{-2mm}}
\foreach \x in {A, ..., Z}{%
    \expandafter\xdef\csname orcid\x\endcsname{\noexpand\href{https://orcid.org/\csname orcidauthor\x\endcsname}{\noexpand\orcidicon}}
}

\begin{document}

\title{Revisiting Continual Semantic Segmentation with Pre-trained Vision Models}

\author{
Duzhen~Zhang$^{*}$,
Yong~Ren$^{*}$,
Wei~Cong,
Junhao~Zheng,
Qiaoyi~Su,
Shuncheng~Jia,
Zhong-Zhi~Li,
Xuanle~Zhao,
Ye~Bai,
Feilong~Chen,
Qi~Tian\orcidA{},~\IEEEmembership{Fellow,~IEEE},
Tielin~Zhang$^{\dagger}$
\thanks{Version: v1 (major update on July 31, 2025)}
\thanks{$^{*}$Equal contribution. $^\dagger$Corresponding author.}
\thanks{Duzhen Zhang is with the Mohamed bin Zayed University of Artificial Intelligence, Abu Dhabi, UAE (E-mail: duzhen.zhang@mbzuai.ac.ae).}
\thanks{Yong Ren, Qiaoyi Su, Shuncheng Jia, Zhong-Zhi Li, Xuanle Zhao, and Ye Bai are with the Institute of Automation, Chinese Academy of Sciences, Beijing, China (E-mail: thurenyong@gmail.com; suqiaoyi2020@ia.ac.cn; jiashuncheng2020@ia.ac.cn; lizhongzhi2022@ia.ac.cn; zhaoxuanle2022@ia.ac.cn, baiye@cau.edu.cn).}
\thanks{Wei Cong is with the Shenyang Institute of Automation, Chinese Academy of Sciences, Shenyang, China (E-mail: congwei45@gmail.com).}
\thanks{Junhao Zheng is with the South China University of Technology, Guangzhou, China (E-mail: junhaozheng47@outlook.com).}
\thanks{Feilong Chen and Qi Tian are with the Huawei Inc., Shenzhen, Guangdong, China (E-mail: phellon.chen@gmail.com; tian.qi1@huawei.com).}
\thanks{Tielin Zhang is with the Center for Excellence in Brain Science and Intelligence Technology, Chinese Academy of Sciences, Shanghai, China (E-mail: zhangtielin@ion.ac.cn).}
}

\markboth{Journal of \LaTeX\ Class Files, January 2025}%
{Shell \MakeLowercase{\textit{et al.}}: A Sample Article Using IEEEtran.cls for IEEE Journals}


\IEEEtitleabstractindextext{

\begin{abstract}

Continual Semantic Segmentation (CSS) seeks to incrementally learn to segment novel classes while preserving knowledge of previously encountered ones. 
Recent advancements in CSS have been largely driven by the adoption of Pre-trained Vision Models (PVMs) as backbones. 
Among existing strategies, Direct Fine-Tuning (DFT)---which sequentially fine-tunes the model across classes---remains the most straightforward approach. 
Prior work often regards DFT as a performance lower bound due to its presumed vulnerability to severe catastrophic forgetting, leading to the development of numerous complex mitigation techniques. 
However, we contend that this prevailing assumption is flawed. 
In this paper, we systematically revisit forgetting in DFT across two standard benchmarks---Pascal VOC 2012 and ADE20K---under eight CSS settings using two representative PVM backbones: ResNet101 and Swin-B. 
Through a detailed probing analysis, our findings reveal that existing methods significantly underestimate the inherent anti-forgetting capabilities of PVMs. 
Even under DFT, PVMs retain previously learned knowledge with minimal forgetting. 
Further investigation of the feature space indicates that the observed forgetting primarily arises from the classifier's drift away from the PVM, rather than from degradation of the backbone representations. 
Based on this insight, we propose DFT*, a simple yet effective enhancement to DFT that incorporates strategies such as freezing the PVM backbone and previously learned classifiers, as well as pre-allocating future classifiers. 
Extensive experiments show that DFT* consistently achieves competitive or superior performance compared to sixteen state-of-the-art CSS methods, while requiring substantially fewer trainable parameters (0.000271\%–0.033194\%) and less training time (30.3\%–45.6\%). 
The code is available at \url{https://github.com/BladeDancer957/RevisitingCSS}.

\end{abstract}

}

\maketitle

\section{Introduction}\label{sec:intro}

\IEEEPARstart{S}{emantic} segmentation has long been recognized as a fundamental task in the Computer Vision (CV) community~\cite{long2015fully, chen2017no, chen2018encoder, guo2018review, garcia2017review}. 
In recent years, Pre-trained Vision Models (PVMs) have achieved remarkable progress across a wide range of downstream CV tasks~\cite{he2016deep, dosovitskiy2020image, liu2021swin, li2022contextual,hou2024conv2former}.
Leveraging PVMs as backbones has become a standard paradigm in semantic segmentation, leading to significant performance gains~\cite{strudel2021segmenter, kirillov2023segment,zhang2022segvit}. 
Nevertheless, conventional semantic segmentation models are predominantly trained and evaluated under static conditions~\cite{geng2020recent}, which diverge from the dynamic and evolving nature of real-world environments~\cite{wang2024get}.
In scenarios involving continuous data streams~\cite{hadsell2020embracing}, models must learn new classes while retaining knowledge of previously seen ones. 
A naive solution to preserving performance is by blending new and old data to retrain the model; however, this approach incurs substantial computational overhead, especially when dealing with large-scale datasets.

To tackle this challenge, Continual Semantic Segmentation (CSS) \cite{ilt, mib, cha2021ssul, plop, lin2023preparing, zhu2023continual,maracani2021recall,dong2023federated} has emerged as a crucial solution, enabling the continuous identification of new classes without requiring complete retraining. 
Among various CSS approaches, Direct Fine-Tuning (DFT) is the most straightforward, involving sequential fine-tuning of the model across learning steps. 
Traditionally, CSS research has regarded DFT as the performance lower bound, assuming it is highly susceptible to catastrophic forgetting \cite{kirkpatrick2017overcoming, mccloskey1989catastrophic,french1999catastrophic, wang2024comprehensive, zheng2025towards, zheng2025lifelong}---a phenomenon where neural networks lose previously acquired knowledge when learning new information. To address this issue, recent studies have introduced various sophisticated methods to mitigate forgetting \cite{yuan2024survey}.

However, we contend that this prevailing assumption may be flawed. 
In this paper, we aim to investigate the following Research Questions (RQs):
\begin{enumerate}[leftmargin=1cm, label=\textbf{RQ\arabic{*}}:]
\item How to measure forgetting in PVMs under DFT?  
\item Do PVMs truly experience forgetting under DFT?  
\item What causes forgetting in DFT?  
\item How to improve DFT with simple strategies?
\end{enumerate}

\begin{figure}[t]
\centering
  \includegraphics[width=1.0\linewidth]{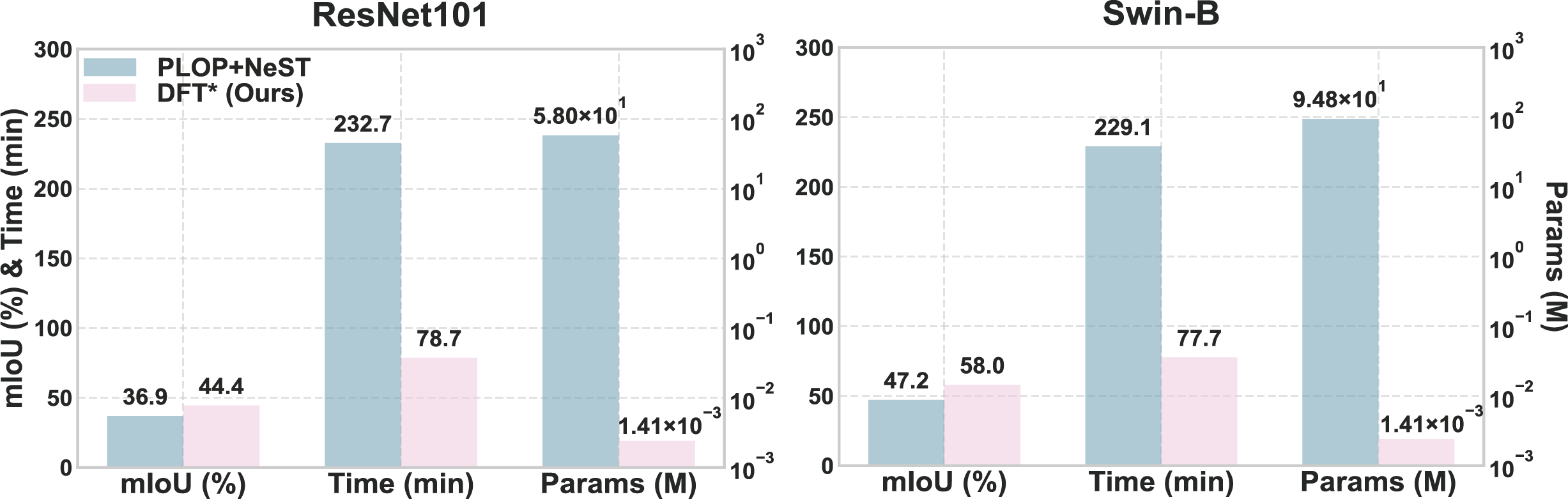}
 \caption{Results on the 10-1 setting of the Pascal VOC 2012 dataset \cite{everingham2015pascal}, using ResNet101 \cite{he2016deep} (trained on a single GPU) and Swin-B \cite{liu2021swin} (trained on two GPUs) as backbones. 
 Our proposed DFT* method surpasses the previous SOTA CSS approach, PLOP+NeST \cite{xie2024early}, in terms of mIoU, while significantly reducing training time and the number of trainable parameters. 
 More detailed results are presented in Section~\ref{sec:exp}.}
\label{fig:intro}
\end{figure}

To answer the above RQs, we conduct extensive experiments to revisit forgetting in DFT on two public datasets (\emph{i.e.}, Pascal VOC 2012 \cite{everingham2015pascal} and ADE20K \cite{zhou2017scene}) under eight CSS settings, using two widely adopted PVM backbones (\emph{i.e.}, ResNet101 \cite{he2016deep} and Swin-B \cite{liu2021swin}). 
Our findings from these experiments are as follows:
\begin{enumerate}

\item From a probing perspective, our findings reveal that existing CSS methods substantially underestimate the intrinsic resilience of PVMs to forgetting. 
Even under DFT, PVMs exhibit strong retention of previously acquired knowledge with minimal performance degradation (see Sections \ref{probing} and \ref{pvms}), particularly in the case of the more capable transformer-based model Swin-B \cite{liu2021swin}.

\item Through feature space analysis, we observe that the forgetting in DFT primarily stems from the classifier weights deviating from the PVM features rather than the actual loss of prior knowledge within the PVM backbone (Section \ref{classifier}).

\item Building on this insight, we propose DFT*, which enhances DFT with simple yet effective strategies, including freezing the PVM backbone and the old classifiers, as well as pre-allocating future classifiers (Section \ref{dft*}).

\item Our DFT* method achieves competitive or superior performance compared to sixteen State-Of-The-Art (SOTA) CSS approaches. 
Notably, DFT* requires only 30.3\%–45.6\% of the training time and 0.000271\%–0.033194\% of the trainable parameters (Section \ref{sec:exp}). 
Results for the 10-1 setting on the Pascal VOC 2012 dataset \cite{everingham2015pascal} are illustrated in Figure \ref{fig:intro}.

\end{enumerate}
Our study calls on the CV community to reconsider and deepen the understanding of forgetting in PVMs.

\section{Preliminaries}\label{sec:preliminary}

\subsection{Problem Formulation}

Following \cite{mib, plop}, CSS aims to learn a model in $t$=$1,...,T$ steps, where a sequential data stream $\{\mathcal{D}^t\}_{t=1}^T$ with corresponding classes $\{\mathcal{C}^t\}_{t=1}^T$ is received. 
At each step $t$, the model is required to learn $\left |\mathcal{C}^t \right |$ new classes, while previous training data is not accessible. 
For an image at the current step, pixels belonging to $\mathcal{C}^t$ are labeled according to their ground truth classes, while the remaining pixels are masked as background. 
After each step $t$, the model should be capable of predicting all classes seen up to that point, $\mathcal{C}^{1:t}$.

At step $t$, the segmentation model consists of a PVM backbone $f^t_{\bm{\theta}}(\cdot)$ and a classifier $g^t_{\bm{\phi}}(\cdot)$, where $\bm{\phi}_t$ represents the newly added classifier weights, and $\bm{\phi}_{1:t-1}$ corresponds to the weights of the classifiers from previous steps.

\subsection{Experimental Setup}

To enable fair comparisons with SOTA CSS approaches, we adopt the experimental setup from \cite{mib, plop, rcil, cong2023gradient, cong2024cs, xie2024early}, including the datasets, CSS settings, evaluation metrics, and foundational implementation details.

\subsubsection{Datasets}

We evaluate on two segmentation datasets: Pascal VOC 2012 \cite{everingham2015pascal}, which includes 20 classes with 10,582 training images and 1,449 validation images, and ADE20K \cite{zhou2017scene}, which consists of 150 classes with 20,210 training images and 2,000 validation images.

\subsubsection{CSS Settings}

In CSS, the training process is divided into $T$ sequential steps, each introducing one or more new classes. 
During step $t$, pixels associated with classes introduced in previous steps are treated as background. 
At evaluation, the model is required to recognize all classes encountered up to the current step. 
Two scenarios are typically considered: Disjoint and Overlapped \cite{mib}. 
The Disjoint scenario assumes that images in the current step contain no pixels from classes to be introduced in future steps. In contrast, the more realistic Overlapped scenario permits the presence of future classes within the current step's images. 
Our method is primarily evaluated under the Overlapped scenario.

The CSS setting is represented as X-Y, where X denotes the number of classes trained in the initial step, and Y indicates the number of classes introduced in subsequent incremental steps. 
We evaluate various CSS settings for each dataset. 
For example, on VOC \cite{everingham2015pascal}, the settings 19-1, 15-5, 15-1, and 10-1 involve learning 19 classes followed by 1 class (2 steps), 15 classes followed by 5 classes (2 steps), 15 classes followed by five 1-class increments (6 steps), and 10 classes followed by ten 1-class increments (11 steps).
The final setting is the most challenging due to its higher number of steps. 
Similarly, for ADE \cite{zhou2017scene}, the settings are 100-50, which involves 100 classes followed by 50 classes (2 steps); 
50-50, which involves 50 classes followed by two 50-class increments (3 steps); 100-10, which involves 100 classes followed by five 10-class increments (6 steps); and 100-5, which involves 100 classes followed by ten 5-class increments (11 steps).

\begin{figure*}[t]
\centering
  \includegraphics[width=1.0\linewidth]{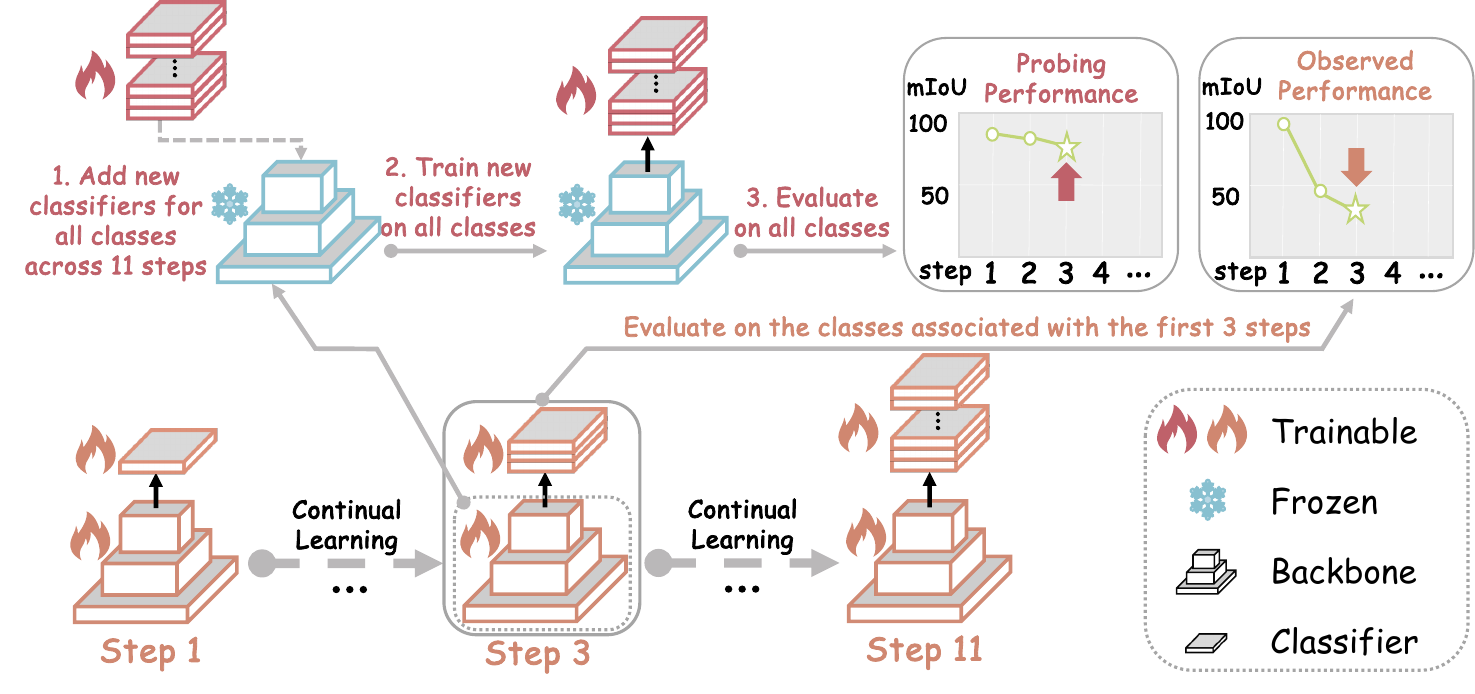}
   \caption{A visual representation of the process for obtaining probing and observed performance after step 3 in a sequence of 11 steps (assume $T$=11). 
   In existing CSS studies, observed performance serves as a measure of forgetting. 
   Probing performance, on the other hand, reveals the extent of forgetting in the PVM backbone, a factor often ignored in prior research.}
\label{fig:prob}
\end{figure*}

\subsubsection{Evaluation Metrics} 

We evaluate the performance of different CSS methods using the traditional metric of mean Intersection over Union (mIoU). 
Specifically, we calculate the mIoU (\%) at the final step $T$ for three categories: the initial classes $\mathcal{C}^1$, the incremental classes $\mathcal{C}^{2:T}$, and the combined set of all classes $\mathcal{C}^{1:T}$ (referred to as ``all''). 
These metrics provide insights into the model's robustness to catastrophic forgetting (model stability), its ability to learn new classes (plasticity), and its overall performance, reflecting the trade-off between these factors. 
Additionally, we track the step-wise performance for all classes encountered up to each step $t$, evaluating $\mathcal{C}^{1:t}$ after each incremental step.

Since the first step is common across all methods, we measure the training time (min) starting from the second step to the final step, maintaining the same operating environment. 
For the trainable parameters metric (M), we calculate the average number of trainable parameters at each step, from the second step through to the last.

\subsubsection{Implementation Details}

We employ PVMs, specifically DeepLabV3 \cite{chen2017deeplab} with ResNet101 \cite{he2016deep} and Swin-B \cite{liu2021swin}, as the backbone for our segmentation model. 
In our experiments, we use SGD as the optimizer for both Pascal VOC 2012 \cite{everingham2015pascal} and ADE20K \cite{zhou2017scene}. 
We also utilize a \emph{poly} schedule for weight decay. 
For the ResNet101 backbone, the model is trained with a batch size of 24 on a single GPU across all experiments. 
For the larger Swin-B backbone, the model is trained with a batch size of 24 on two GPUs. 
Given the commonality of the initial step across all methods, we reuse the weights obtained during this phase. 
All experiments are performed on NVIDIA L40S GPUs with 48 GB of memory. 
Additional implementation details---including data preprocessing, training commands, hyperparameter settings, evaluation protocols, and all runtime shell scripts---are available in the public codebase at \url{https://github.com/BladeDancer957/RevisitingCSS}.

\subsection{Related Work}

Previous CSS work assumes that DFT is highly susceptible to catastrophic forgetting \cite{kirkpatrick2017overcoming, mccloskey1989catastrophic} and has thus introduced various complex methods to address this issue \cite{yang2023label,yin2025beyond,cong2025lightweight}.

For example, MiB~\cite{mib} introduces unbiased cross-entropy and stores the model from the previous step to distill old knowledge into the current model. 
This distillation approach was widely adopted and laid the foundation for subsequent CSS methods. 
PLOP~\cite{plop} uses pseudo-labels to address background shift and employs intermediate features from the PVM backbone to transfer prior knowledge. 
SDR~\cite{sdr} proposes a contrastive learning technique aimed at minimizing intra-class feature distances. 
RCIL~\cite{rcil} integrates reparameterization into CSS with a complementary network structure. 
GSC~\cite{cong2023gradient} investigates CSS by incorporating gradient and semantic compensation. 
EWF~\cite{xiao2023endpoints} combines old and new knowledge through a weight fusion strategy. 
SSUL~\cite{cha2021ssul} uses auxiliary data to train an `unknown' classifier, which is then utilized to initialize new classifiers. 
DKD~\cite{baek2022decomposed} introduces a decomposed knowledge distillation approach, improving model rigidity and stability. 
AWT~\cite{goswami2023attribution} applies gradient-based attribution to transfer the relevant weights from old classifiers to new ones. Cs$^2$K~\cite{cong2024cs} proposes class-specific and class-shared knowledge guidance for CSS. 
NeST~\cite{xie2024early} presents a classifier pre-tuning method that strikes a balance between plasticity and stability.

However, we contend that this assumption is flawed and the above CSS methods significantly underestimate the inherent anti-forgetting capabilities of PVMs. 
From a probing perspective, we observe that even under DFT, PVMs retain knowledge with minimal forgetting. 
The forgetting observed in DFT is mainly due to the deviation of the classifier from the PVMs, rather than the loss of prior knowledge in PVMs. 
Based on this insight, we propose DFT*, combining DFT with a few simple strategies. 
Our results show that DFT* outperforms or matches the performance of previous SOTA CSS methods, while requiring considerably less training time and fewer trainable parameters.

\section{Revisiting the Forgetting in DFT}\label{sec:method}

\subsection{How to Measure Forgetting in PVMs?}\label{probing}

In this section, we explain how to assess forgetting within PVMs under DFT. 
To achieve this, we employ the probing technique, a robust method for examining the capability of backbones to encode and retain information relevant to specific tasks \cite{chenforgetting,taocan,davari2022probing,wu2022pretrained,zheng2024learn}.

As illustrated in Figure \ref{fig:prob}, we assess the knowledge retained in PVMs under DFT across learning steps by introducing a \textbf{probing classifier} after each step $t$. 
These classifiers are placed atop the PVM and trained on \textit{all classes} encountered throughout the learning process. Subsequently, we evaluate both the PVM and the probing classifiers on all classes to determine the \textbf{probing performance}. 
Notably, this evaluation does not interfere with the CSS training process, as the PVM backbone remains frozen while training the probing classifiers. 
For clarity, we refer to the original step-wise CSS performance, measured using the DFT method, as the \textbf{observed performance}. 
Additionally, while the original classifiers only make predictions for classes associated with \textit{learned} steps, the probing classifiers predict classes from \textit{all} steps in CSS.

\begin{figure*}[t]
\centering
  \includegraphics[width=1.0\linewidth]{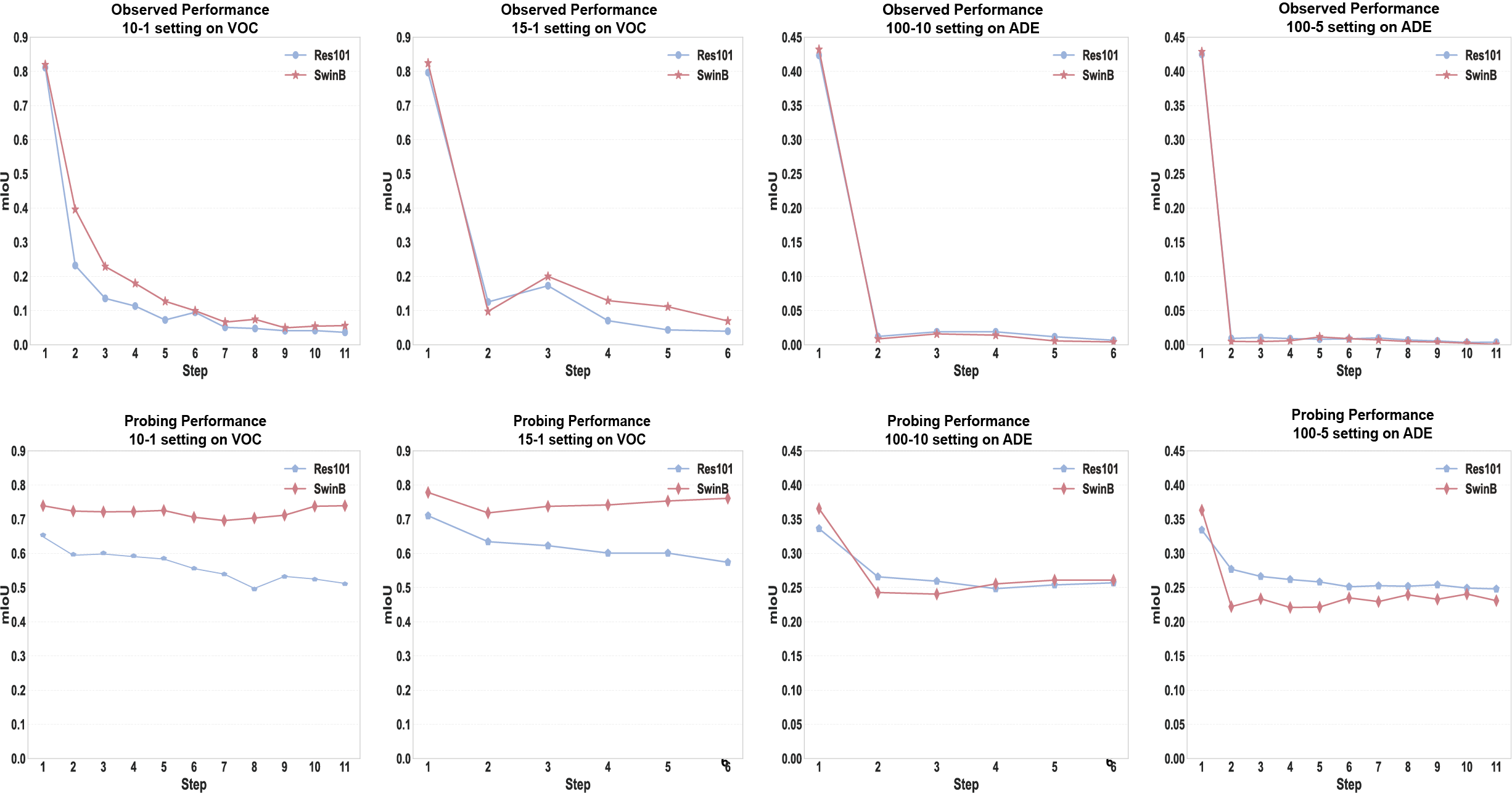}
\caption{Observed and probing performance under the 10-1 and 15-1 settings of Pascal VOC 2012 dataset \cite{everingham2015pascal}, and the 100-10 and 100-5 settings of ADE20K dataset \cite{zhou2017scene}, using ResNet101 \cite{he2016deep} and Swin-B \cite{liu2021swin} as PVM backbones. 
The first row shows a sharp drop in observed performance after the first step, which remains consistently low as more classes are introduced---reflecting the typical pattern of catastrophic forgetting under DFT. 
In contrast, the second row shows that probing performance remains high and largely stable after the initial step.}
\label{fig:obs_prob}
\end{figure*}

\begin{figure*}[t!]
\centering
  \includegraphics[width=1.0\linewidth]{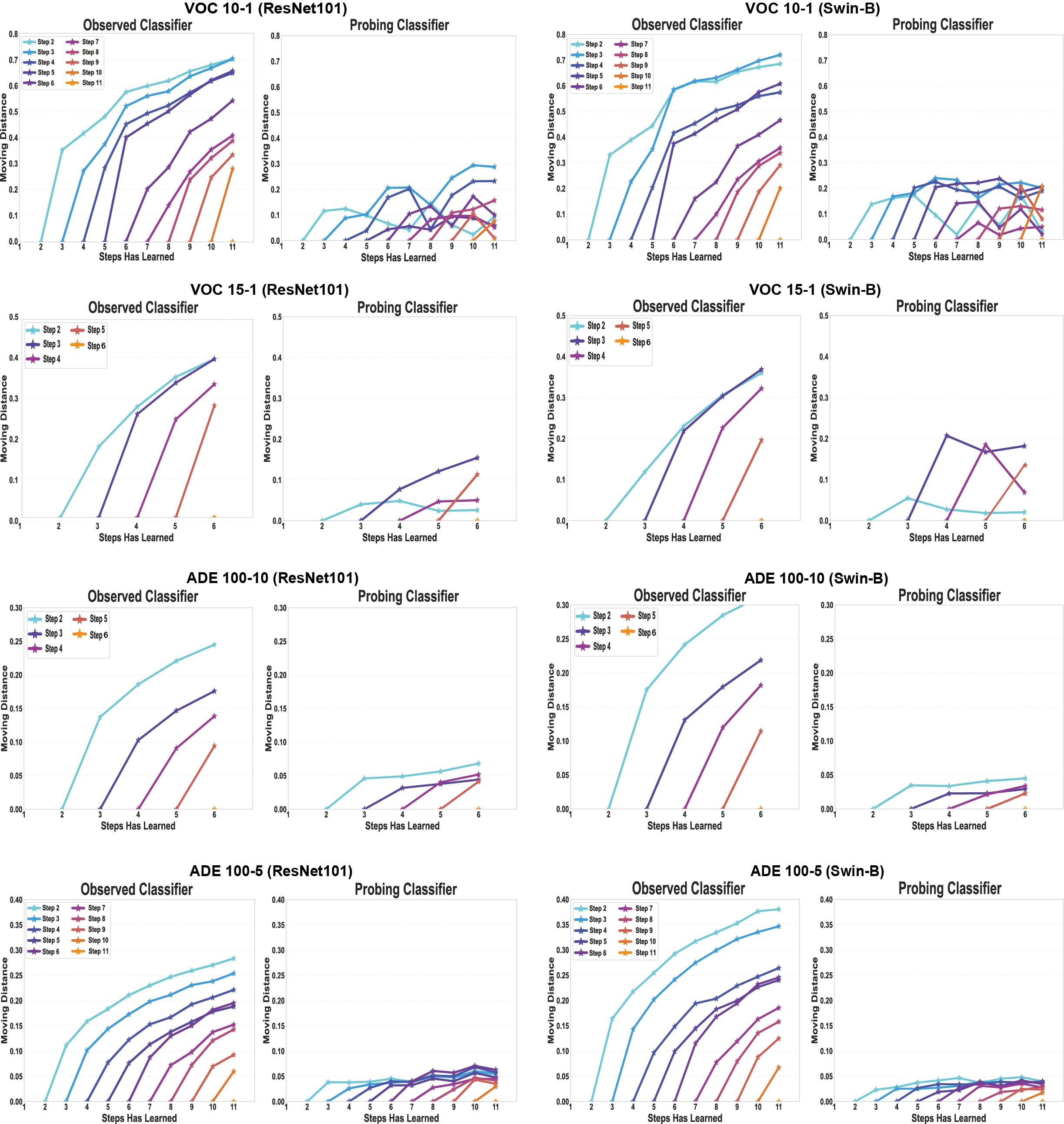}
\caption{Comparison of the moving distance of class weights between the observed classifier and the probing classifier under DFT, evaluated on two CSS settings of the Pascal VOC 2012 dataset \cite{everingham2015pascal} (10-1 and 15-1) and two CSS settings of the ADE20K dataset \cite{zhou2017scene} (100-10 and 100-5), using ResNet101 \cite{he2016deep} and Swin-B \cite{liu2021swin} as PVM backbones. 
The class weights of observed classifiers undergo substantial shifts compared to those of probing classifiers, suggesting that forgetting occurs as the original class weights are displaced from their initial and optimal positions.}
\label{fig:moving}
\end{figure*}

\subsection{Do PVMs Experience Forgetting under DFT?}\label{pvms}

Building on the previously defined concepts of observed and probing performance, we present a comparative analysis of these metrics across two CSS settings on the Pascal VOC 2012 dataset (10-1 and 15-1) \cite{everingham2015pascal}, and two settings on the ADE20K dataset (100-10 and 100-5) \cite{zhou2017scene}, using ResNet101 \cite{he2016deep} and Swin-B \cite{liu2021swin} as the PVM backbones. The results are visualized in Figure \ref{fig:obs_prob}.

The first row of Figure \ref{fig:obs_prob} illustrates a sharp decline in observed performance immediately after the first step, followed by persistently low performance throughout the subsequent steps, aligning with the well-established pattern of catastrophic forgetting under the DFT method.
In contrast, the second row reveals a markedly different trend. 
Overall, the PVMs achieve high probing performance after the initial step, and this performance remains largely stable throughout subsequent steps.

Specifically, under the 10-1 and 15-1 settings of Pascal VOC 2012, the transformer-based Swin-B \cite{liu2021swin} consistently maintains high and stable probing performance across all continual learning steps. 
While ResNet101 \cite{he2016deep} exhibits a slight performance drop after the first step, it still achieves relatively high and stable probing results in later steps. 
The robustness of Swin-B can be attributed to its transformer architecture \cite{vaswani2017attention}, which provides stronger pretraining capabilities and richer feature representations.

On the ADE20K dataset \cite{zhou2017scene}, under the 100-10 and 100-5 settings, both Swin-B and ResNet101 show a modest decline in performance after the first step, but continue to maintain overall high and stable probing performance across steps. These results suggest that, even when sequentially adapting to new classes under DFT, PVMs effectively retain their ability to recognize all classes throughout the CSS process.

In sum, these findings indicate that previous research has largely underestimated the inherent resistance of PVMs to forgetting. 
Even under DFT, PVMs retain knowledge with minimal degradation. 
This raises a crucial question: What is the primary cause of forgetting in the DFT method?

\subsection{What Causes Forgetting in DFT?}\label{classifier}

As discussed in the previous section, PVM backbones exhibit minimal degradation in probing performance under DFT, suggesting that they retain knowledge with little forgetting. 
This implies that the primary cause of forgetting in DFT stems from the classifier. 
In this section, we delve deeper into how forgetting occurs within classifiers via feature space analysis.

Previous studies \cite{wu2019large, hou2019learning, zheng2024balancing,zheng2024learn} have identified a class imbalance issue between old and new classes in continual learning, particularly under DFT. 
Specifically, the logits of new classes tend to be significantly larger than those of old classes. 
Since the magnitude of logits is influenced by both class weights (\emph{i.e.}, the row vectors of the classifier's weight matrix) and the features extracted from the PVM backbone, we infer that catastrophic forgetting \textbf{primarily} stems from the cosine similarity between class weights and feature representations.

\begin{figure}[t!]
\centering
  \includegraphics[width=1.0\linewidth]{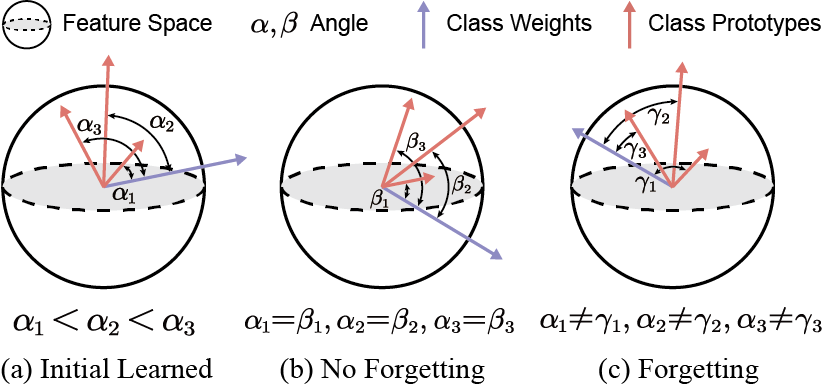}
\caption{Illustration of the moving distance between class weights and class prototypes. (a) Depicts the cosine similarity between class weights and prototypes after learning a new step. (b) Demonstrates that forgetting does not occur when the moving distance (\emph{i.e.}, relative cosine similarity) remains unchanged. (c) Highlights that forgetting takes place when the moving distance shifts.}
\label{fig:feature_space}
\end{figure}

To quantify this effect, we introduce \textbf{Moving Distance (MD)}, a metric that measures changes in the cosine similarity between class weights and feature representations extracted from the PVM backbone during DFT. Rather than considering all features, we compute class prototypes, which represent the average feature embeddings of pixels belonging to the same class. A visual depiction of this concept is provided in Figure \ref{fig:feature_space}. After each learning step, the alignment between class weights and class prototypes is optimal for the newly learned classes. Forgetting occurs only when this relative positioning shifts over time. Ideally, if the angular relationship between class weights and class prototypes remains unchanged, forgetting would not take place. In practice, the smaller the variation in this angle---reflected by a lower MD---the less forgetting is observed.

We formally define the MD of class weights of the $t$-th step ($t$=2,$\dots$,$T$) at the $t$+$k$-th step ($k$=1,$\dots$,$T$-$t$) as follows:
\begin{equation}
\footnotesize
    MD_{t+k}^{t} = \frac{1}{mn}\sum_{m=1}^{|\mathcal{C}^{all}|}\sum_{n=1}^{|\mathcal{C}^{t}|}|\mathbf{Cos}_{t+k}^{t}[m,n] - \mathbf{Cos}_{t}^{t}[m,n]|\text{,}
\end{equation}
where $\mathcal{C}^{all}$ and $\mathcal{C}^{t}$ denote the class sets across all $T$ steps and at the $t$-th step, respectively. 
$\mathbf{Cos}$ refers to the cosine similarity matrix calculated between all pairs of class prototypes and class weights. 
Specifically, $\mathbf{Cos}_{t+k}^{t}$ is the cosine similarity matrix computed between class prototypes measured at step $t$+$k$ and class weights at step $t$. 
The element $\mathbf{Cos}_{t+k}^{t}[m,n]$ corresponds to the cosine similarity at position $[m,n]$ in this matrix, representing the similarity between the $m$-th class prototype of all classes and the $n$-th class weight of classes at step $t$.

We summarize the MD results under DFT for two CSS settings on the Pascal VOC 2012 dataset \cite{everingham2015pascal} (10-1 and 15-1) and two CSS settings on the ADE20K dataset \cite{zhou2017scene} (100-10 and 100-5), using ResNet101 \cite{he2016deep} and Swin-B \cite{liu2021swin} as PVM backbones. 
These results are illustrated in Figure \ref{fig:moving}. 
The findings reveal that the weights of old classes in the observed classifiers drift significantly from their original positions, leading to forgetting. 
In contrast, the weights of old classes in the probing classifier remain relatively close to their initial positions. 
This analysis suggests that forgetting in DFT primarily stems from deviations of the classifier weights from the PVM representations.

\subsection{How to Improve DFT with Simple Strategies?}\label{dft*}

\begin{figure}[t]
\centering
  \includegraphics[width=1.0\linewidth]{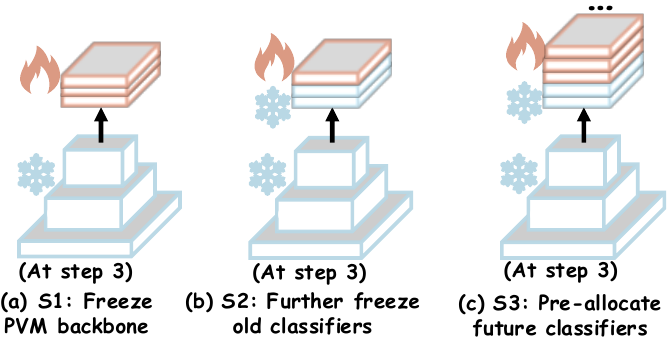}
\caption{An illustration of the proposed DFT*, combining DFT with three straightforward strategies.}
\label{fig:strategy}
\end{figure}

In this section, we introduce DFT*, which is inspired by the insights gained from the forgetting observed in DFT. 
We refer to this enhanced version as DFT*, and an illustration of the strategy is shown in Figure \ref{fig:strategy}. 
The previous section presents the following key insights regarding DFT:
\begin{enumerate}[leftmargin=1cm, label=\textbf{I\arabic{*}}:]
\item The PVM backbone does not acquire new knowledge during CSS under DFT.

\item The PVM backbone delivers the high probing performance once adapted to downstream steps, with minimal performance loss when learning on additional new steps (shown in the second row of the Figure \ref{fig:obs_prob}).

\item The forgetting in DFT primarily results from the classifier weights deviating from the PVM features (shown in Figure \ref{fig:moving}).

\end{enumerate}

Consequently, we propose a few simple strategies, along with their rationale, to bridge the gap between the probing and observed performance in DFT:
\begin{enumerate}[leftmargin=1cm, label=\textbf{S\arabic{*}}:]
\item Freeze the PVM backbone, referred to as \textbf{FixB}, as recommended by \textbf{I1} and \textbf{I2}.

\item Freeze both the PVM backbone and the old classifiers, referred to as \textbf{FixBC}, as proposed by \textbf{I3}. Only the new classifier is updated, while the old classifiers and the PVM backbone remain fixed. \textbf{S2} ensures that the relative position of class weights in the classifier remains consistent with the class prototypes extracted from the PVM backbone, addressing the issue highlighted in \textbf{I3}. Additionally, \textbf{S1} can help alleviate this problem from the PVM backbone side.

\item Pre-allocate future classifiers, termed \textbf{FixBC+P}. For instance, when learning classes in the third step of the 10-1 setting (11 steps) on VOC \cite{everingham2015pascal}, the classifiers for steps 4 to 11 are pre-allocated. These future classifiers are trained in advance during the third step, \textbf{even if no instances from steps 4 to 11 are available}. This method enhances the forward compatibility of the classifiers, making it easier for new classifiers to adapt to new classes when old ones are frozen \cite{zhou2022forward}. \textbf{S3} requires prior knowledge of the total number of steps, but this can be bypassed by consistently maintaining a classifier with a class count that exceeds the number of encountered classes.

\end{enumerate}
The above strategies are implemented starting from the second learning step, while the first step is the same as the previous CSS methods and can share the same base model.

\section{Comparing DFT* with SOTA CSS Methods}\label{sec:exp}

\begin{table*}[t]
\centering
\caption{The mIoU (\%) on the final step of the Pascal VOC 2012 dataset \cite{everingham2015pascal} under four CSS settings. \textcolor{deepred}{\textbf{Red}} indicates the highest performance. Baseline results are cited from \cite{xie2024early}, \cite{cong2024cs}, and \cite{cong2023gradient}. $^{\ddagger}$ denotes results from our re-implementation using the official code. Our DFT* method achieves competitive or superior performance compared to sixteen SOTA CSS methods.}
    \label{tab:voc}
    \resizebox{1.0\textwidth}{!}{
    \begin{tabular}{c|c|cc|c|cc|c|cc|c|cc|c}
        \toprule[1.2pt]
        \multirow{2}{*}{\textbf{Method}} &\multirow{2}{*}{\textbf{Backbone}}& \multicolumn{3}{c|}{\textbf{10-1(11 steps)}} & \multicolumn{3}{c|}{\textbf{15-1(6 steps)}} & \multicolumn{3}{c|}{\textbf{15-5(2 steps)}} & \multicolumn{3}{c}{\textbf{19-1(2 steps)}} \\
        & &  0-10 & 11-20 & all & 0-15 & 16-20 & all & 0-15 & 16-20 & all & 0-19 & 20 & all \\
         
        \hline
        Joint & Res101 & 78.8 &77.7 & 78.3 & 79.8 & 73.4 & 78.3 & 79.8 & 73.4 & 78.3 & 78.2 & 80.0 & 78.3 \\      
        \hline  
        EWC \cite{kirkpatrick2017overcoming} (PNAS'17)  & Res101 &  6.6 &  4.9 &  5.8 &  0.3 &  4.3 &  1.3 & 24.3 & 35.5 & 27.1 & 26.9 & 14.0 & 26.3 \\      
        
        LwF \cite{li2017learning} (TPAMI'17)            & Res101 &  8.0 &  2.0 &  4.8 &  6.0 &  3.9 &  5.5 & 58.9 & 36.6 & 53.3 & 51.2 &  8.5 & 49.1 \\ 

        LwF-MC \cite{rebuffi2017icarl} (CVPR'17)        & Res101 & 11.2 &  2.5 &  7.1 &  6.4 &  8.4 &  6.9 & 58.1 & 35.0 & 52.3 & 64.4 & 13.3 & 61.9 \\ 
        
        ILT~\cite{ilt} (ICCV'19)                        & Res101 &  7.2 &  3.7 &  5.5 &  9.6 &  7.8 &  9.2 & 67.8 & 40.6 & 61.3 & 68.2 & 12.3 & 65.5 \\  

        MiB~\cite{mib} (CVPR'20)                        & Res101 & 12.2 & 13.1 & 12.6 & 38.0 & 13.5 & 32.2 & 76.4 & 49.4 & 70.0 & 71.2 & 22.1 & 68.9 \\  

        SDR~\cite{sdr} (CVPR'21)                        & Res101 & 32.4 & 17.1 & 25.1 & 44.7 & 21.8 & 39.2 & 75.4 & 52.6 & 69.9 & 69.1 & 32.6 & 67.4 \\ 

        MiB+AWT~\cite{goswami2023attribution} (WACV'23) & Res101 & 33.2 & 18.0 & 26.0 & 59.1 & 17.2 & 49.1 & 77.3 & 52.9 & 71.5 & -   & -    & -    \\ 

        PLOP~\cite{plop} (CVPR'21)                      & Res101 & 44.0 & 15.5 & 30.5 & 65.1 & 21.1 & 54.6 & 75.7 & 51.7 & 70.1 & 75.4 & 37.4 & 73.5 \\  

        PLOP+UCD~\cite{yang2022uncertainty} (TPAMI'22)  & Res101 & 42.3 & 28.3 & 35.3 & 66.3 & 21.6 & 55.1 & 75.0 & 51.8 & 69.2 & 75.9 & 39.5 & 74.0 \\   

        RCIL~\cite{rcil} (CVPR'22)                      & Res101 & 55.4 & 15.1 & 34.3 & 70.6 & 23.7 & 59.4 & 78.8 & 52.0 & 72.4 & 77.0 & 31.5 & 74.7 \\   

        GSC~\cite{cong2023gradient} (TMM'23)            & Res101 & 50.6 & 17.3 & 34.7 & 72.1 & 24.4 & 60.8 & 78.3 & 54.2 & 72.6 & 76.9 & 42.7 & 75.3 \\  

        SPPA~\cite{sppa} (ECCV'22)                      & Res101 & -    & -    & -    & 66.2 & 23.3 & 56.0 & 78.1 & 52.9 & 72.1 & 76.5 & 36.2 & 74.6 \\  

        ALIFE~\cite{oh2022alife} (NeurIPS'22)           & Res101 & -    & -    & -    & 64.4 & 34.9 & 57.4 & 77.2 & 52.5 & 71.3 & 76.6 & 49.4 & 75.3 \\  

       MiB+EWF \cite{xiao2023endpoints} (CVPR'23)       & Res101 & 56.0 & 16.7 & 37.3 & 78.0 & 25.5 & 65.5 & -    & -    & -    & 77.8 & 12.2 & 74.7 \\  

        MiB+Cs$^2$K \cite{cong2024cs} (ECCV'24)         & Res101 & 43.0 & 35.2 & 39.3 & 76.2 & 41.8 & \textcolor{deepred}{\textbf{68.0}} & -    & -    & -    & -    & -    & -    \\ 

        MiB+NeST \cite{xie2024early} (ECCV'24)          & Res101 & 52.3 & 21.0 & 37.4 & 61.7 & 20.4 & 51.8 & 77.1 & 50.1 & 70.7 & 71.7 & 28.2 & 69.7 \\

        PLOP+NeST$^{\ddagger}$ \cite{xie2024early} (ECCV'24)  & Res101 & 53.4 & 17.9 & 36.5 & 72.3 & 28.3 & 61.8 & 77.2 & 50.2 & 70.8 & 76.7 & 31.4 & 74.5 \\ 
        
        PLOP+NeST \cite{xie2024early} (ECCV'24)         & Res101 & 54.2 & 17.8 & 36.9 & 72.2 & 33.7 & 63.1 & 77.6 & 55.8 & 72.4 & 77.0 & 49.1 & \textcolor{deepred}{\textbf{75.7}} \\

        RCIL+NeST \cite{xie2024early} (ECCV'24)         & Res101 & 51.4 & 20.9 & 36.8 & 71.9 & 28.0 & 61.4 & 79.0 & 52.8 & \textcolor{deepred}{\textbf{72.8}} & 77.0 & 33.3 & 74.9 \\ 
        \hline

        DFT                                              & Res101 &  6.4 &  0.8 &  3.7 &  4.6 &  2.0 &  4.0 &  6.5 & 33.4 & 12.9 & 12.8 & 11.6 & 12.8 \\    

        \rowcolor{LightBlue} DFT* (DFT+FixB) (Ours)      & Res101 & 27.7 &  3.9 & 16.4 & 53.3 & 17.7 & 44.8 & 48.4 & 37.9 & 45.9 & 63.8 & 31.6 & 62.3 \\   

        \rowcolor{LightBlue} DFT* (DFT+FixBC) (Ours)     & Res101 & 34.2 & 31.5 & 32.9 & 64.9 & 26.1 & 55.7 & 70.2 & 32.9 & 61.4 & 78.1 & 17.8 & 75.2 \\

        \rowcolor{LightBlue} DFT* (DFT+FixBC+P) (Ours)   & Res101 & 51.7 & 36.3 & \textcolor{deepred}{\textbf{44.4}} & 69.7 & 28.3 & 59.8 & 72.9 & 37.5 & 64.5 & 77.5 & 27.6 & 75.2 \\

       \hline\hline
       Joint                        & Swin-B & 80.4 &79.7 &80.1 &81.1 &76.7 &80.1 &81.1 &76.7& 80.1 &80.0 &80.7 &80.1\\ 
       \hline  
        MiB~\cite{mib} (CVPR'20)                        & Swin-B & 11.4 & 18.9 & 15.0 & 35.0 & 43.2 & 36.9 & 80.7 & 66.5 & 77.3 & 79.2 & 60.2 & 78.3 \\ 
        
       MiB+NeST \cite{xie2024early} (ECCV'24)           & Swin-B & 65.2 & 35.8 & 51.2 & 77.0 & 53.3 & 71.4 & 81.2 & 67.4 & 77.9 & 79.7 & 60.0 & 78.8 \\ 
       
        PLOP \cite{plop} (CVPR'21)                      & Swin-B & 37.8 & 23.1 & 30.8 & 74.1 & 52.1 & 68.9 & 80.1 & 68.1 & 77.2 & 77.0 & 65.8 & 76.4 \\ 

        PLOP+NeST$^{\ddagger}$ \cite{xie2024early} (ECCV'24)  & Swin-B & 58.1 & 31.4 & 45.4 & 76.8 & 55.9 & 71.8 & 81.9 & 68.5 & \textcolor{deepred}{\textbf{78.7}} & 79.5 & 68.1 & 79.0 \\
        
        PLOP+NeST \cite{xie2024early} (ECCV'24)         & Swin-B & 64.3 & 28.3 & 47.2 & 76.8 & 57.2 & 72.2 & 80.5 & 70.8 & 78.2 & 79.6 & 70.2 & 79.1 \\

        \hline 

        DFT                                              & Swin-B &  6.3 &  5.7 &  6.0 & 10.2 & 12.8 & 10.8 & 18.3 & 54.4 & 26.9 & 24.8 & 51.9 & 26.1 \\   

        \rowcolor{LightBlue} DFT* (DFT+FixB) (Ours)      & Swin-B & 26.0 & 6.0 & 16.5  & 56.8 & 39.7 & 52.7 & 49.7 & 53.6 & 50.6 & 68.9 & 45.7 & 67.8 \\

        \rowcolor{LightBlue} DFT* (DFT+FixBC) (Ours)     & Swin-B & 54.8 & 48.9 & 52.0 & 78.5 & 48.3 & 71.3 & 79.3 & 50.3 & 72.4 & 80.9 & 39.1 & 78.9 \\

        \rowcolor{LightBlue} DFT* (DFT+FixBC+P) (Ours)   & Swin-B & 62.5 & 53.0 & \textcolor{deepred}{\textbf{58.0}} & 80.2 & 54.4 & \textcolor{deepred}{\textbf{74.1}} & 81.1 & 57.7 & 75.5 & 80.7 & 51.9 & \textcolor{deepred}{\textbf{79.4}} \\

        \bottomrule[1.2pt]
    \end{tabular}
    }
\end{table*}

\begin{figure*}[t]
\centering
  \includegraphics[width=1.0\linewidth]{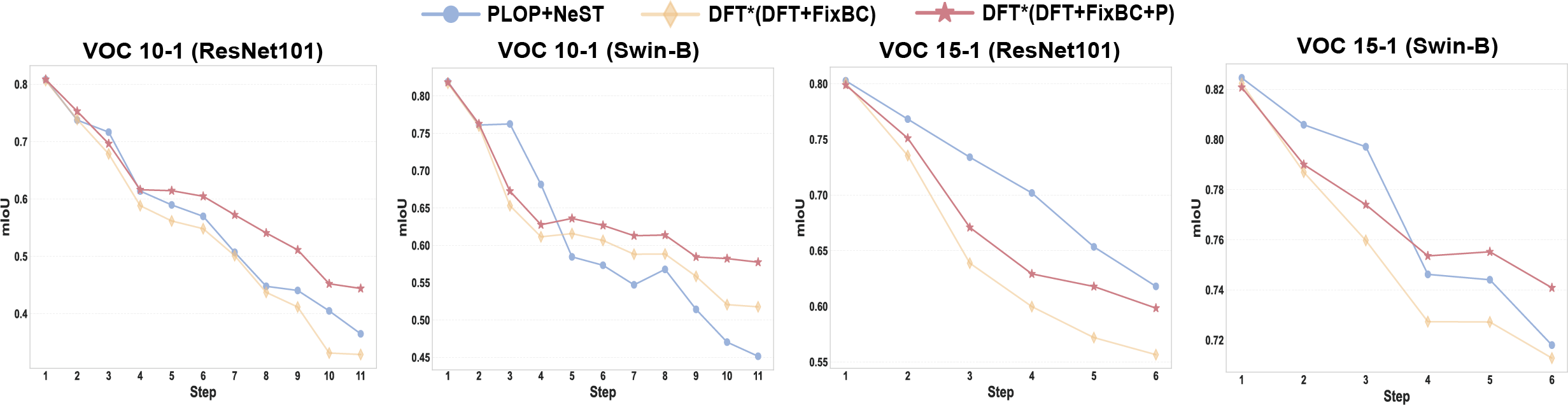}
\caption{Step-wise comparison of the SOTA CSS method PLOP+NeST \cite{xie2024early} with our proposed DFT* variants (DFT+FixBC and DFT+FixBC+P) on the Pascal VOC 2012 dataset \cite{everingham2015pascal}, under the 10-1 and 15-1 settings, using ResNet101 \cite{he2016deep} and Swin-B \cite{liu2021swin} as PVM backbones. 
Our DFT* approach consistently achieves competitive or superior performance at each step compared to PLOP+NeST \cite{xie2024early} (results reproduced using the official code under the same experimental environment).}
\label{fig:voc_step}
\end{figure*}

\subsection{Results on VOC Dataset}

\subsubsection{CSS Performance}

As shown in Table \ref{tab:voc}, we compare DFT* with sixteen SOTA CSS methods across four different settings---10-1, 15-1, 15-5, and 19-1---on the Pascal VOC 2012 dataset \cite{everingham2015pascal}, using two PVM backbones: ResNet101 \cite{he2016deep} and Swin-B \cite{liu2021swin}. 
Despite its simplicity, DFT* delivers competitive or even superior performance across all settings. 
In particular, under the challenging 10-1 setting, which involves the largest number of learning steps, DFT* (DFT+FixBC+P) significantly outperforms previous SOTA methods. Furthermore, DFT* consistently achieves large improvements over the original DFT baseline. We emphasize that we \textbf{do not claim DFT* achieves SOTA performance in every setting}. Instead, our goal is to demonstrate that \textbf{DFT* provides a strong and competitive baseline across most CSS scenarios}, and should therefore be considered in future CSS research.

Among the three DFT* variants, DFT* (DFT+FixB) mitigates the primary cause of forgetting---deviation between the classifier and the PVM backbone---by addressing it from the backbone side, leading to moderate performance gains over the original DFT. 
DFT* (DFT+FixBC), which freezes both the old classifiers and the PVM backbone, further reduces this deviation and achieves a substantial improvement. 
Finally, DFT* (DFT+FixBC+P) pre-allocates future classifiers, enhancing their forward compatibility and enabling new classifiers to adapt more effectively to novel classes when old classifiers are frozen, resulting in additional performance gains. 
Moreover, DFT* performs better with the Swin-B backbone \cite{liu2021swin} than with the ResNet101 backbone \cite{he2016deep}, as this freezing-based approach is particularly well-suited for stronger PVMs that encapsulate richer pre-trained knowledge. 

As illustrated in Figure \ref{fig:voc_step}, our DFT* approach---including the DFT+FixBC and the DFT+FixBC+P variants---consistently delivers competitive or superior performance at every step compared to the previous SOTA CSS method, PLOP+NeST \cite{xie2024early}.

\begin{figure*}[t]
\centering
  \includegraphics[width=1.0\linewidth]{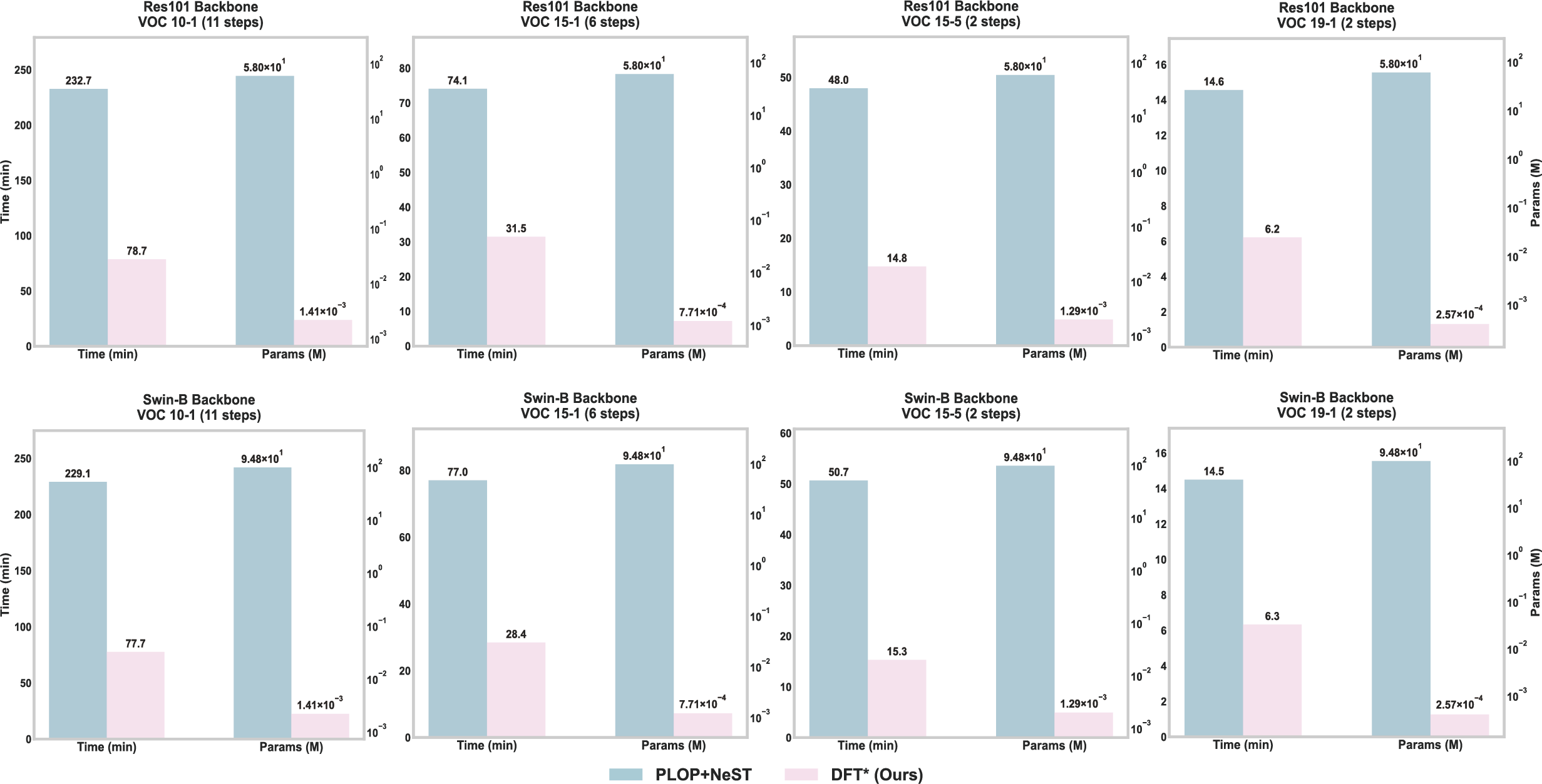}
\caption{Comparison of training time and trainable parameters between our DFT* (DFT+FixBC+P) and the SOTA CSS method PLOP+NeST \cite{xie2024early} (results reproduced using the official code under the same experimental environment) across four settings (10-1, 15-1, 15-5, and 19-1) on the Pascal VOC 2012 dataset \cite{everingham2015pascal}, using ResNet101 \cite{he2016deep} (trained on single GPU) and Swin-B \cite{liu2021swin} (trained on two GPUs) as backbones. DFT+FixBC+P significantly reduces both the number of trainable parameters and training time.}
\label{fig:voc_time}
\end{figure*}

\subsubsection{Training Time \& Trainable Parameters}

As shown in Figure \ref{fig:voc_time}, we present a comparison of training time and the number of trainable parameters between our proposed DFT* (DFT+FixBC+P) and the SOTA CSS method PLOP+NeST \cite{xie2024early}. 
The results are reproduced using the official implementation of PLOP+NeST under the same experimental environment to ensure fairness. 
The comparison is conducted across four settings---10-1, 15-1, 15-5, and 19-1---on the Pascal VOC 2012 dataset \cite{everingham2015pascal}, employing two PVM backbones: ResNet101 \cite{he2016deep} and Swin-B \cite{liu2021swin}.

Across all settings and backbones, DFT* (DFT+FixBC+P) demonstrates a clear advantage in efficiency, requiring substantially less training time and dramatically fewer trainable parameters than PLOP+NeST. 
Specifically, DFT* (DFT+FixBC+P) achieves only about 30.3\%–43.7\% of the training time and an extremely small fraction---0.000271\%–0.002436\% of the trainable parameters---compared to PLOP+NeST. 
Notably, ResNet101 experiments are conducted on a single GPU, whereas Swin-B experiments utilize two GPUs, which can occasionally result in shorter training times for Swin-B despite its more complex architecture.

These results highlight the lightweight nature of DFT*, which minimizes optimization overhead by freezing key components, while still maintaining competitive or superior CSS performance. 
Such efficiency makes DFT* attractive for resource-constrained scenarios, providing a practical and high-performing baseline for future CSS research.

\begin{table*}[t]
\caption{The mIoU (\%) on the final step of the ADE20K dataset \cite{zhou2017scene} under four CSS settings. \textcolor{deepred}{\textbf{Red}} indicates the highest performance. Baseline results are cited from \cite{xie2024early} and \cite{cong2024cs}. $^{\ddagger}$ denotes results from our re-implementation using the official code. Our DFT* method achieves competitive or superior performance compared to twelve SOTA CSS methods.}
    \label{tab:ade}
    \centering
    \resizebox{1.0\textwidth}{!}{
    \begin{tabular}{c|c|cc|c|cc|c|cc|c|cc|c}
        \toprule[1.2pt]
        \multirow{2}{*}{\textbf{Method}} & \multirow{2}{*}{\textbf{Backbone}} & \multicolumn{3}{c|}{\textbf{100-50(2 steps)}} & \multicolumn{3}{c|}{\textbf{100-10(6 steps)}} & \multicolumn{3}{c}{\textbf{100-5(11 steps)}} & \multicolumn{3}{c}{\textbf{50-50(3 steps)}}\\
        & & 0-100 & 101-150 & all & 0-100 & 101-150 & all & 0-100 & 101-150 & all &0-50&51-150&all \\
        \hline 
Joint& Res101& 44.3 &28.2& 38.9 &44.3 &28.2& 38.9 &44.3& 28.2 &38.9 & 51.1 & 33.3 & 38.9\\

   \hline 
        ILT~\cite{ilt} (ICCV'19)                        & Res101 & 18.3 & 14.8 & 17.0 &  0.1 &  2.9 &  1.1 &  0.1 &  1.3 &  0.5 &13.6& 6.2 &9.7 \\

        MiB~\cite{mib} (CVPR'20)                        & Res101 & 40.5 & 17.2 & 32.8 & 38.2 & 11.1 & 29.2 & 36.0 &  5.6 & 25.9 & 45.6 & 21.0 & 29.3 \\

        MiB+AWT~\cite{goswami2023attribution} (WACV'23) & Res101 & 40.9 & 24.7 & 35.6 & 39.1 & 21.3 & 33.2 & 38.6 & 16.0 & 31.1 & 46.6 & 26.9 & 33.5 \\

        PLOP~\cite{plop} (CVPR'21)                      & Res101 & 41.9 & 14.9 & 32.9 & 40.5 & 13.6 & 31.6 & 39.1 & 7.8 & 28.7 & 48.8 & 21.0 & 30.4\\ 

        PLOP+UCD~\cite{yang2022uncertainty} (TPAMI'22)  & Res101 & 42.1 & 15.8 & 33.3 & 40.8 & 15.2 & 32.3 & -    & -   & -   & 47.1 & 24.1 & 31.8  \\ 

        RCIL~\cite{rcil} (CVPR'22)                      & Res101 & 42.3 & 18.8 & 34.5 & 39.3 & 17.6 & 32.1 & 38.5 & 11.5 & 29.6 & 48.3 & 25.0 & 32.5\\

        GSC~\cite{cong2023gradient} (TMM'23)            & Res101 & 42.4 & 19.2 & 34.8 & 40.8 & 17.6 & 32.6 & 39.5 & 11.2 & 30.2 & 46.2 & 26.2 & 33.0\\

        SPPA~\cite{sppa} (ECCV'22)                      & Res101 & 42.9 & 19.9 & 35.2 & 41.0 & 12.5 & 31.5 & -    & -    & -  & 49.8 & 23.9 & 32.5   \\

        ALIFE~\cite{oh2022alife} (NeurIPS'22)           & Res101 & 42.2 & 23.1 & 35.9 & 41.0 & 22.8 & \textcolor{deepred}{\textbf{35.0}} & -    & -    & -  & 49.0 & 25.7 & 33.6   \\

        MiB+EWF \cite{xiao2023endpoints} (CVPR'23)      & Res101 & 41.2 & 21.3 & 34.6 & 41.5 & 16.3 & 33.2 & 41.4 & 13.4 & 32.1  & --&--&-- \\ 

        MiB+Cs$^2$K \cite{cong2024cs} (ECCV'24)         & Res101 & -    & -    & -    & 42.4 & 17.2 & 34.1 & 41.4 & 13.4 & 32.1 & --&--&--  \\

        MiB+NeST \cite{xie2024early}  (ECCV'24)         & Res101 & 40.3 & 24.6 & 35.1 & 40.2 & 20.6 & 33.7 & 39.9 & 18.0 & \textcolor{deepred}{\textbf{32.7}} & 45.6 & 26.8 & 33.2 \\ 
        
        PLOP+NeST$^{\ddagger}$ \cite{xie2024early} (ECCV'24)  & Res101 & 42.3 &  22.2 & 35.6 & 39.7 &  19.7 & 33.1 & 37.9 &  14.9 & 30.3 & 48.5 & 26.1 & 33.7 \\
    
        PLOP+NeST \cite{xie2024early} (ECCV'24)         & Res101 & 42.2 & 24.3 & \textcolor{deepred}{\textbf{36.3}} & 40.9 & 22.0 & 34.7 & 39.3 & 17.4 & 32.0  & 48.7 & 27.7 & \textcolor{deepred}{\textbf{34.8}}  \\
        
        RCIL+NeST \cite{xie2024early} (ECCV'24)         & Res101 & 42.3 & 22.8 & 35.8 & 40.7 & 19.0 & 33.5 & 39.4 & 15.5 & 31.5 & 48.2 & 27.4 & 34.4 \\ 
        
        \hline 
        DFT                                              & Res101 &  0.1 & 13.3 & 4.5 &  0.1 &  2.3 &  0.8 &  0.1 &  0.9 &  0.3 & 0.2& 7.6&  5.1\\   
        
        \rowcolor{LightBlue} DFT* (DFT+FixBC) (Ours)     & Res101 & 42.3 & 19.1 & 34.6 & 41.9 & 16.7 & 33.5 & 40.8 & 14.8 & 32.2 & 48.5 & 21.3 & 30.5 \\  
        \rowcolor{LightBlue} DFT* (DFT+FixBC+P) (Ours)   & Res101 & 41.7 & 19.3 & 34.3 & 41.2 & 18.5 & 33.7 & 40.6 & 15.4 & 32.3 & 48.3 & 22.8 & 31.4 \\

        \hline\hline
        Joint & Swin-B & 43.4 & 31.9 & 39.6 & 43.4 & 31.9 & 39.6 & 43.4 & 31.9 & 39.6 & 50.7 &33.9& 39.6 \\
        \hline

        MiB \cite{mib} (CVPR'20)                        & Swin-B & 42.7 & 26.1 & 37.2 & 40.2 & 15.0 & 31.8 & 39.1 &  8.6 & 29.0  & 48.3 & 26.8 & 34.1\\ 
        
        MiB+NeST \cite{xie2024early} (ECCV'24)          & Swin-B & 42.8 & 27.8 & 37.9 & 41.8 & 23.8 & 35.9 & 40.5 & 19.9 & 33.7 & 49.7 & 29.3 & 36.2 \\ 
        
        PLOP \cite{plop} (CVPR'21)                      & Swin-B & 43.4 & 17.1 & 34.7 & 41.4 & 17.7 & 33.6 & 39.7 & 13.6 & 31.0 & 50.5 & 24.1 &33.0 \\ 
        
        PLOP+NeST$^{\ddagger}$ \cite{xie2024early} (ECCV'24)  & Swin-B & 43.4 & 27.4 & \textcolor{deepred}{\textbf{38.1}} & 40.8 &  24.0 & 35.2 & 38.3 & 20.9 & 32.6 & 50.5 & 29.9 & \textcolor{deepred}{\textbf{36.9}} \\

        PLOP+NeST \cite{xie2024early} (ECCV'24)         & Swin-B & 43.5 & 26.5 & 37.9 & 41.7 & 24.2 & 35.9 & 39.7 & 18.3 & 32.6 & 50.6 & 28.9 & 36.2 \\ 
        
        \hline  
        DFT                                              & Swin-B &  0.1 &  8.1 &  2.8 &  0.1 &  1.2 &  0.4 &  4.5 &  0.1 &  0.1 &  0.2 & 6.9& 4.6\\   
        
        \rowcolor{LightBlue} DFT* (DFT+FixBC) (Ours)     & Swin-B & 43.1 & 25.2 & 37.1 & 42.6 & 24.4 & 36.6 & 41.8 & 21.4 & 35.0 & 49.7 & 26.6 & 34.4 \\
        \rowcolor{LightBlue} DFT* (DFT+FixBC+P) (Ours)   & Swin-B & 43.5 & 26.4 & 37.8 & 43.2 & 24.1 & \textcolor{deepred}{\textbf{36.9}} & 42.6 & 21.7 & \textcolor{deepred}{\textbf{35.7}} & 50.2 & 27.9 & 35.4 \\ 
        
        \bottomrule[1.2pt]
    \end{tabular}
}
\end{table*}

\begin{figure*}[t]
\centering
  \includegraphics[width=1.0\linewidth]{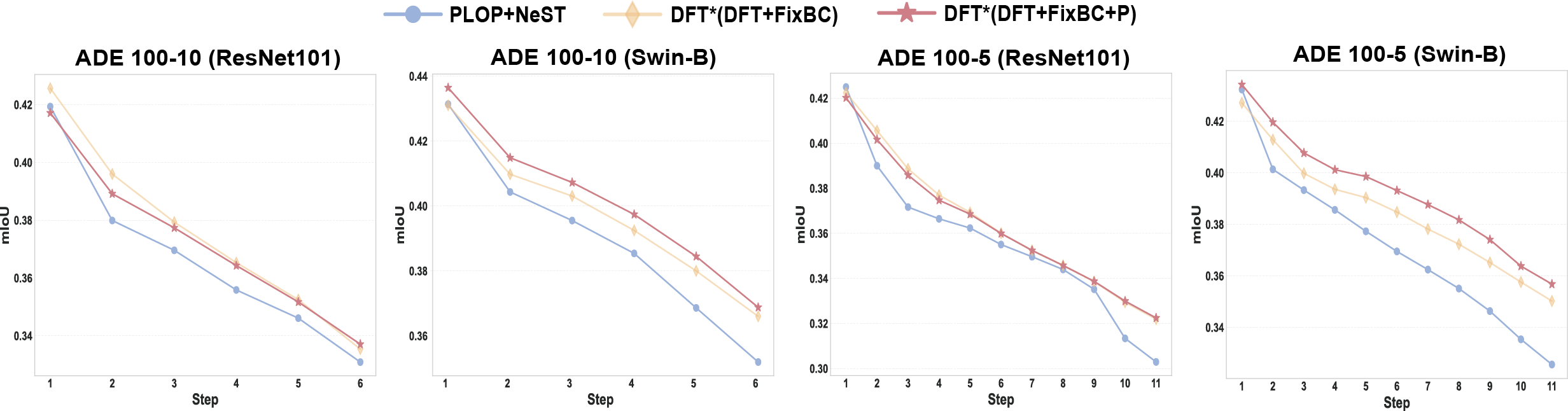}
\caption{Step-wise comparison of the SOTA CSS method PLOP+NeST \cite{xie2024early} with our proposed DFT* variants (DFT+FixBC and DFT+FixBC+P) on the ADE20K dataset \cite{zhou2017scene}, under the 100-10 and 100-5 settings, using ResNet101 \cite{he2016deep} and Swin-B \cite{liu2021swin} as PVM backbones. 
Our DFT* approach consistently achieves competitive or superior performance at each step compared to PLOP+NeST \cite{xie2024early} (results reproduced using the official code under the same experimental environment).}
\label{fig:ade_step}
\end{figure*}

\subsection{Results on ADE Dataset}

\subsubsection{CSS Performance}

As shown in Table \ref{tab:ade}, we compare DFT* against twelve SOTA CSS methods across four different settings---100-50, 100-10, 100-5, and 50-50---on the ADE20K dataset \cite{zhou2017scene}, using two PVM backbones: ResNet101 \cite{he2016deep} and Swin-B \cite{liu2021swin}. 
Despite its simplicity, DFT* delivers competitive or even superior performance across all settings. 
Notably, in the challenging 100-5 setting, which involves the largest number of learning steps, DFT* (DFT+FixBC+P) with the Swin-B backbone establishes a new SOTA performance level. 
Furthermore, DFT* consistently achieves substantial improvements over the original DFT across all experimental settings. We emphasize that we \textbf{do not claim DFT* to be the absolute SOTA method in every scenario}. 
Instead, our goal is to demonstrate that \textbf{DFT* provides a strong and competitive baseline in almost all CSS settings}, making it a valuable reference point for future research in CSS.

Consistent with the results on the VOC dataset \cite{everingham2015pascal}, DFT* achieves better performance with the Swin-B \cite{liu2021swin} than with the ResNet101 \cite{he2016deep}. However, the performance of the two DFT* variants, DFT+FixBC and DFT+FixBC+P, remains remarkably close across all four settings. 
Furthermore, as shown in Figure \ref{fig:ade_step}, our DFT* approach---encompassing the DFT+FixBC and DFT+FixBC+P variants---consistently outperforms the previous SOTA CSS method, PLOP+NeST \cite{xie2024early}, delivering superior performance at every step under both the 100-10 and 100-5 settings of the ADE20K dataset, using both ResNet101 and Swin-B backbones.

\begin{figure*}[t]
\centering
  \includegraphics[width=1.0\linewidth]{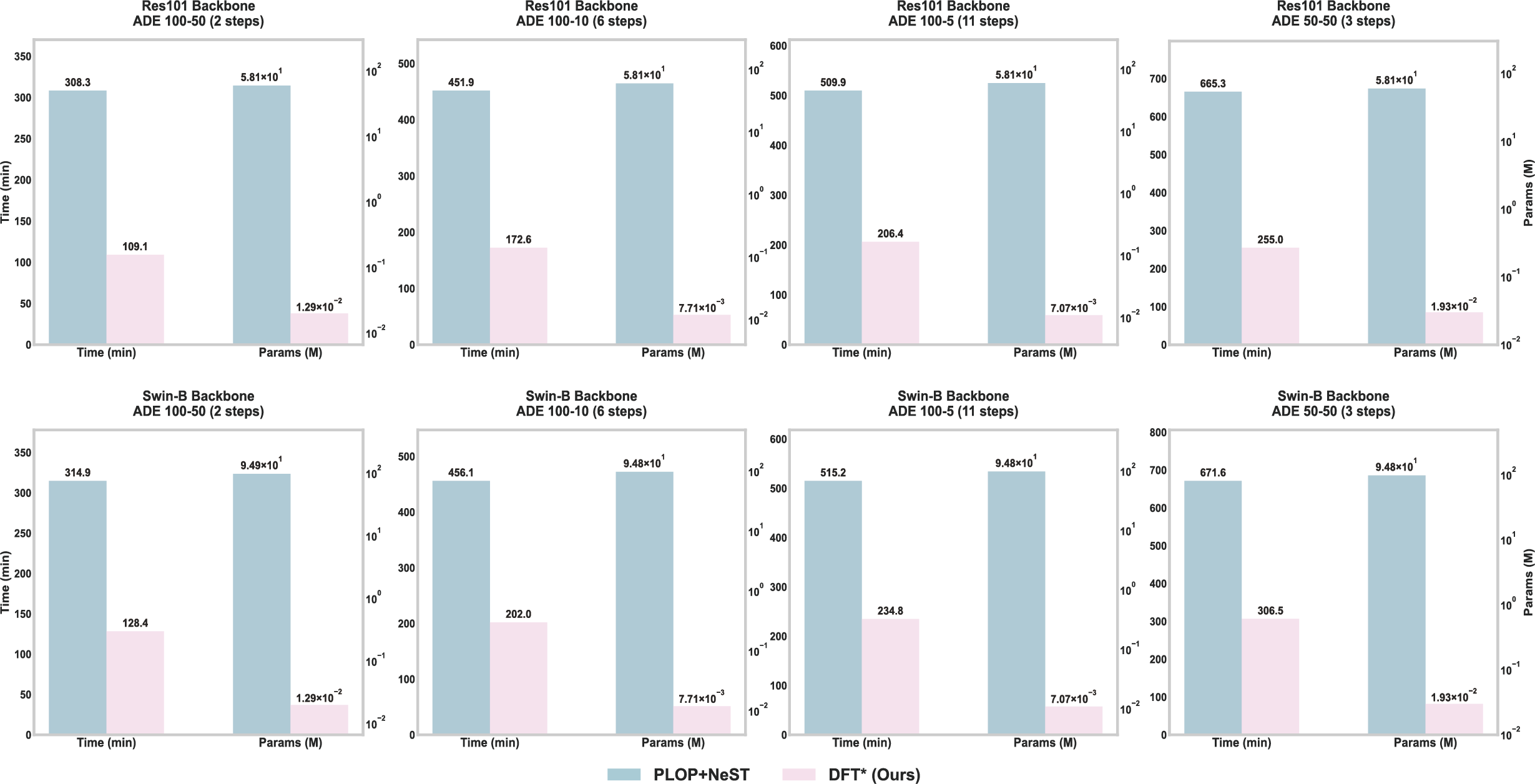}
\caption{Comparison of training time and trainable parameters between our DFT* (DFT+FixBC+P) and the SOTA CSS method PLOP+NeST \cite{xie2024early} (results reproduced using the official code under the same experimental environment) across four settings (100-50, 100-10, 100-5, and 50-50) on the ADE20K dataset \cite{zhou2017scene}, using ResNet101 \cite{he2016deep} (trained on single GPU) and Swin-B \cite{liu2021swin} (trained on two GPUs) as backbones. DFT+FixBC+P significantly reduces both the number of trainable parameters and training time.}
\label{fig:ade_time}
\end{figure*}

\subsubsection{Training Time \& Trainable Parameters}

As shown in Figure \ref{fig:ade_time}, we compare the training time and the number of trainable parameters between our proposed DFT* (DFT+FixBC+P) and the SOTA CSS method PLOP+NeST \cite{xie2024early}. 
The results for PLOP+NeST are reproduced using its official implementation under the same experimental conditions to ensure a fair comparison. 
The comparison spans four settings---100-50, 100-10, 100-5, and 50-50---on the ADE20K dataset \cite{zhou2017scene}, using two PVM backbones: ResNet101 \cite{he2016deep} and Swin-B \cite{liu2021swin}.

In all settings and with both backbones, DFT* (DFT+FixBC+P) clearly outperforms PLOP+NeST in terms of efficiency, requiring significantly less training time and far fewer trainable parameters. Specifically, DFT* (DFT+FixBC+P) consumes only about 35.4\%–45.6\% of the training time and an extremely small fraction---0.007452\%–0.033194\%---of the trainable parameters compared to PLOP+NeST.

These results underscore the lightweight nature of DFT*, which minimizes optimization overhead by freezing key components, while still maintaining competitive or superior CSS performance. 
This efficiency makes DFT* particularly well-suited for large-scale or resource-constrained environments, offering a practical and high-performing baseline for future research in CSS.

\section{Conclusion}\label{sec:conclusion}

In this paper, we revisited the role of DFT in CSS and challenged the conventional belief that it inherently suffers from catastrophic forgetting. 
Through extensive analysis across multiple datasets, CSS settings, and PVM backbones, we demonstrated that PVMs possess strong anti-forgetting properties, which existing studies have largely underestimated. 
Our probing and feature space analysis revealed that forgetting in DFT is not primarily due to the degradation of learned representations but rather the misalignment between the classifier's weights and the features extracted from the PVM backbone. 
Building upon this insight, we introduced DFT*, a simple yet effective approach that leverages the intrinsic strengths of PVMs for CSS. 
Experimental results demonstrated that DFT* consistently delivers competitive or even superior performance compared to over ten SOTA CSS methods. Importantly, it achieves this with significantly reduced training time and fewer trainable parameters. These findings highlight the need to reassess prior assumptions in CSS research and advocate for more efficient, frozen-based approaches in future studies.

\bibliography{reference}
\bibliographystyle{IEEEtran}

\end{document}